\documentclass[letterpaper, 10 pt, conference]{ieeeconf}  

\IEEEoverridecommandlockouts                              

\usepackage{graphics} 
\usepackage{epsfig} 
\usepackage{mathptmx} 
\usepackage{times} 
\usepackage{amsmath} 
\usepackage{amssymb}  
\usepackage{algorithm}
\usepackage{algorithmic}
\usepackage{booktabs}
\usepackage{multirow}
\usepackage{siunitx}
\usepackage{subcaption}
\usepackage{comment}
\usepackage{xcolor}
\usepackage[colorinlistoftodos,textwidth=1.22cm]{todonotes}

\title{\LARGE \bf Online Multimodal Transportation Planning using Deep Reinforcement Learning
}

\author{Amirreza Farahani$^{1}$, Laura Genga$^{1}$ and Remco Dijkman$^{1}$
\thanks{*The work leading up to this paper is partly funded by the European Commission under the FENIX project (grant nr. INEA/CEF/TRAN/M2018/1793401).}
\thanks{$^{1}$Amirreza Farahani, Laura Genga and Remco Dijkman are with the Department  of Industrial Engineering, Eindhoven University of Technology, Eindhoven 5612 AZ, The Netherlands
        {\tt\small email: \{a.farahani, l.genga, r.m.dijkman\}@tue.nl}}%
}

\usepackage{fancyhdr}
\begin{document}

\maketitle
\thispagestyle{fancy}
\pagestyle{empty}

\begin{abstract}

In this paper we propose a Deep Reinforcement Learning approach to solve a multimodal transportation planning problem, in which containers must be assigned to a truck or to trains that will transport them to their destination. While traditional planning methods work \lq\lq offline" (i.e., they take decisions for a batch of containers before the transportation starts), the proposed approach is \lq\lq online", in that it can take decisions for individual containers, while transportation is being executed. Planning transportation online helps to effectively respond to unforeseen events that may affect the original transportation plan, thus supporting companies in lowering transportation costs. We implemented different container selection heuristics within the proposed Deep Reinforcement Learning algorithm and we evaluated its performance for each heuristic using data that simulate a realistic scenario, designed on the basis of a real case study at a logistics company. The experimental results revealed that the proposed method was able to learn effective patterns of container assignment. It outperformed tested competitors in terms of total transportation costs and utilization of train capacity by 20.48\% to 55.32\% for the cost and by 7.51\% to 20.54\% for the capacity. Furthermore, it obtained
results within 2.7\% for the cost and 0.72\% for the capacity of the optimal solution generated by an Integer Linear Programming solver in an offline setting.


\end{abstract}

\section{INTRODUCTION}

This paper introduces an online planning algorithm that we developed for a logistics company, based on Deep Reinforcement Learning (DRL). One of the crucial challenges faced by the company and other companies like it, is multimodal transportation planning, in which a container must be assigned to a transportation resource for onward transportation.

\begin{figure}
\begin{center}
    
\includegraphics[scale=0.2]{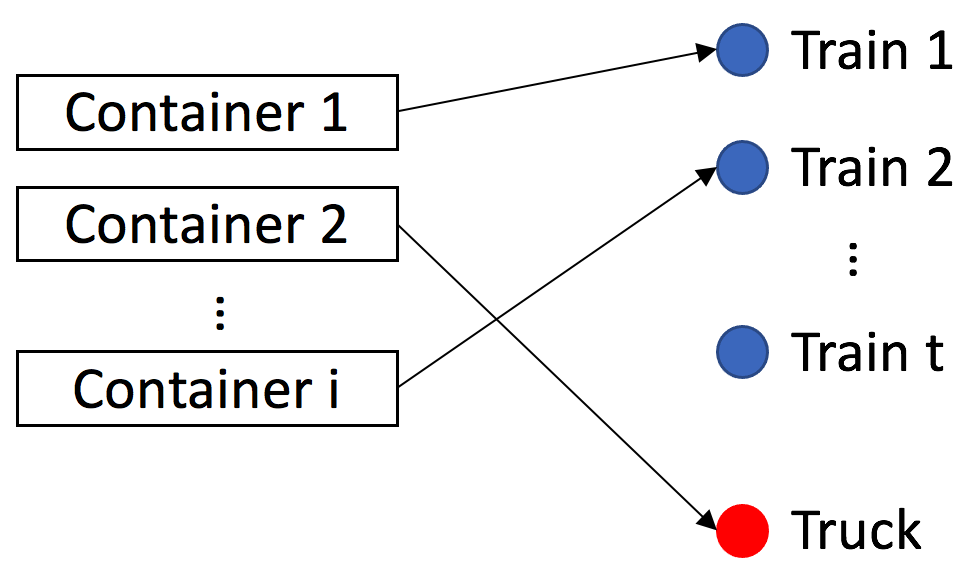}
\caption{Planning problem}
\label{fig:planning_problem}
\end{center}
\vspace{-0.9cm}
\end{figure}

Fig.~\ref{fig:planning_problem} shows a simple example of a multimodal transportation planning problem. Given a set of containers, each with its arrival time and due date at destination, and a set of available vehicle options, each with its transportation costs, and arrival/departure time, the goal is to assign each container to one of the available options, in such a way that the total cost is minimized. As trains have a lower cost than trucks, solving the planning problem corresponds to allocating as many containers to trains as possible.

Multimodal transportation 
involves complex operational processes, each including many operational activities in different organizations such as seaports, airports, logistics companies, train stations, etc. Any delays, operational errors, and so on, in these operational activities can lead to unexpected events such as, for instance, delays in train arrivals  or containers arrivals. 
At the same time, new containers continuously arrive and customers may require unexpected changes to their orders. 
These 
unexpected events introduce unpredictable dynamicity in multimodal transportation,
which has a strong impact on the extent to which the plan can be executed. Consequently, transportation planners are continuously replanning, while the plan is being executed.
Traditional offline planning methods cannot help transportation planners with that task, because these methods assign batch of containers to available vehicles in one go. However, 
when an unforeseen event happens, the planner needs to replan only a single container or the few containers
that are affected by such event. We refer to this as online (re)planning.
Currently, there is little support for online planning, usually carried out by means of heuristics whose outputs may be far from the optimal. 
To fix this gap, in this work 
we propose an online planning method for multimodal transportation based on DRL. The method is able to learn rules to assign individual containers to available vehicles, while the plan is being executed. To the best of our knowledge, such an approach has not yet been studied in this domain.

We tested our method using data simulating a realistic scenario, designed on the basis of a real case study of multimodal transportation planning at a logistics company. 
Our results show that our algorithm can learn rules to effectively assign containers to trains and trucks, one container at a time. Furthermore, it outperformed the tested competitors in terms of total transportation costs and utilization of train capacity, generating a solution close to the offline optimal one.


The rest of this paper is organized as follows. Section \ref{section2} introduces relevant related literature.
Section \ref{section3} provides a formal definition of the multimodal transportation problem. Section \ref{section4} introduces our method.
Section \ref{section5} discusses our experiments and evaluation results. Finally, Section \ref{section6} draws conclusions and delineates future works.

\section{Literature Review}\label{section2}

In this section, first we discuss approaches related to transportation planning in presence of dynamicity. Then, we discuss Deep Reinforcement Learning (DRL) applications in the operation research area.



\subsection{Multimodal Freight Transportation Planning}
Table~\ref{table1} categorizes
related work on multimodal transportation planning with dynamicity on the basis of the following features: (1) Modality considered (rail/road/air/water), (2) Planning type (offline/online), (3) Planning method, (4) Dynamicity, where ``general'' stands for methods able to deal with any unexpected event, and ``specific'' stands for methods considering only specific types of unexpected events, e.g., the train arrival time.

\begin{table*}
\begin{center}
\caption{Methods for multimodal transportation planning in the presence of unexpected events}\label{table1}
\setlength\tabcolsep{1.5pt}
\begin{tabular}{l l l l l l}
\hline
\textbf{Reference} &\quad  \textbf{Modality} &\quad  \textbf{Planning Type} &\quad  \textbf{Methods} &\quad  \textbf{Dynamicity}\\
\hline

~\cite{Ishfaq2012,Yang2016} & \quad Rail, road, air & \quad  Offline & \quad  Model-driven  & \quad  Specific \\
~\cite{Brands2014,Sun2018,Sun2019}& \quad  Rail, road & \quad Offline & \quad  Model-driven  & \quad  Specific  \\
~\cite{VanRiessen2016} & \quad Rail, road, water & \quad Online/Offline & \quad  Data-driven  & \quad  General \\
~\cite{Demir2015,Demir2016,Tian2017} & \quad   Rail, road, air, water & \quad   Offline & \quad  Model-driven  & \quad  Specific  \\
~\cite{Rivera2017,QU2019} & \quad  Rail, road, water & \quad Offline & \quad  Model-driven  & \quad  Specific  \\
~\cite{Gumuskaya2020} & \quad  Road, water & \quad Offline & \quad  Model-driven  & \quad  Specific \\
\hline
\end{tabular}
\end{center}
\vspace{-0.5cm}
\end{table*}

Most of the methods in Table~\ref{table1} are offline~\cite{Ishfaq2012,Brands2014,Demir2015,Demir2016,Tian2017,Yang2016,Rivera2017,Sun2018,Sun2019,QU2019,Gumuskaya2020} meaning they assume to have a-priori information about all containers, and plan all containers together. 
Still, these papers take dynamicity into account to some extent.
However, in most cases this is realized by assuming the shipments to have a known probability distribution, then integrating the unexpected events via offline model-driven approaches methods such as stochastic programming or robust optimization.
An exception is~\cite{VanRiessen2016}, where they mix data-driven approaches with model-driven stochastic methods. 
This method is also able to do replanning after any unexpected events for each container separately. 
To this end, it analyzes the solution structure of a centralized optimization method, which uses offline analysis and classification on historic data to derive online decision-making rules for suitable allocations of containers to inland services. The weak point of this approach is the offline learning phase. 
This model needs to re-train periodically to learn new patterns. 
In contrast, our method is able to replan each container individually after any unexpected and can learn patterns and rules of transportation online. 



\subsection{Deep Reinforcement Learning Applications in Decision Making and Combinatorial Optimization Problems}


One popular application of Reinforcement Learning in transportation optimization area is the Vehicle Routing Problem (VRP). The objective of VRP is minimizing the total route length while satisfying the demand from all customers~\cite{Nazari2018}, which is modeled as a Traveling Salesman Problem (TSP) optimization problem. A number of Deep Reinforcement Learning have been proposed to tackle this problem. For example,
~\cite{Applegate2006} introduced a transportation Pointer Networks (PtrNet)~\cite{Vinyals2017} able to learn a sequence model coupled with an attention mechanism trained to output TSP tours. A few years later, in~\cite{Bello2017}, to train a DNN they use a policy gradient and a variant of the Asynchronous Advantage Actor-Critic (A3C) method, proposed by~\cite{Mnih2016}. The authors of~\cite{Nazari2018}use the Asynchronous Actor-Critic Agents (A3C) algorithm which is provided by~\cite{Mnih2016}. In~\cite{Khalil2017}, a structure2vec (S2V) model was trained to output the ordering of partial tours using Deep Q-Learning (DQN). One year after that, ~\cite{Deudon2018} proposed a hybrid approach using 2-opt local search on top of tours trained via Policy Gradient. In~\cite{Kool2019}, they extended network consideration using a reinforce method with a greedy rollout baseline. In other recent works, the authors of~\cite{daCosta2020} propose a Deep Reinforcement Learning algorithm trained using Policy Gradient to learn improvement heuristics based on 2-opt moves for the TSP and in~\cite{Zhao2020} they use a hybrid of Deep Reinforcement Learning and local search for the VRP.

Another relevant domain where DRL is often applied, is manufacturing. In~\cite{Waschneck2018}, they introduce a Deep Reinforcement Learning method in a dynamic manufacturing environment to allocate waiting jobs to available machines/resources. They apply the DQN algorithm and have an agent for each work center. In~\cite{Chen2019}, authors propose a Reinforcement Learning based Assigning Policy (RLAP)  method for multi-projects scheduling in cloud manufacturing to minimize both the total makespan and the logistical distance. To minimize the makespan for an MCP scheduling problem, the authors of~\cite{Park2019} propose a reinforcement learning (RL) algorithm to setup a change scheduling method. ~\cite{Paraschos2020} discusses a Reinforcement Learning method to find the optimal trade-off between conflicting performance metrics for the optimization of the total expected profit of the system. 

Applications of RL methods in decision making and Combinatorial Optimization Problems (COP) are not limited to VRP or manufacturing area.
In~\cite{Afshar2019-1}, RL is used for deriving the optimal ordering of a network in real time bidding systems. In ~\cite{Afshar2019-2}, a deep Reinforcement Learning approach using state aggregation is developed for solving knapsack problems in the business field. In the supply-chain domain,~\cite{Meisheri2020} authors use a Reinforcement Learning method for a large variable-dimensional inventory management problem, while~\cite{Ghavamipoor2019} proposes a Reinforcement Learning based model for adaptive service quality management in E-Commerce websites. 
This analysis shows that there are other Reinforcement Learning applications for various optimization problems.
However, to the best of our knowledge, no previous work proposed to exploit Deep Reinforcement Learning in multimodal transportation planning.




\section{Problem Definition}\label{section3}
In this section, we provide a formal definition for the multimodal transportation planning problem we aim to solve. We provide two formulations. First, we define the problem in an offline setting, as a  classical combinatorial optimization problem. 
Then, we define the problem in an online setting in the form of a Markov Decision Process (MDP). 

\subsection{Offline Planning Problem Definition} \label{section3.1}

We represent the offline multimodal transportation planning problem as a mathematical programming problem. Variables used to model the problem are described in Table~\ref{table2}. Given a multimodal transportation problem with containers and transportation resources with their schedule and capacity, the goal is to determine an assignment of containers to available vehicles, such that the total cost of transportation is minimal. Formally, this is expressed by Equation \ref{equ1}. This minimization problem has to fulfill the following constraints: (1) each container is assigned to exactly one train or truck 
(Equation \ref{equ3}),(2) a container should be planned on a train only if the train departs on or after the earliest availability day of the container (Equation \ref{equ4}), (3) containers planned on a train should arrive at the latest on their latest arrival day (Equation \ref{equ5}), (4) the maximum capacity of a train can not be exceeded (Equation \ref{equ6}). Consequently, the transportation problem can be defined as:

\begin{table}
\begin{center}
\caption{List of Integer Linear Programming Elements}\label{table2}
\setlength\tabcolsep{3pt}
\begin{tabular}{ l l}
\hline
\textbf{Sets} &  \\
\(I\) & set of containers \\
\(T\) & set of trains \\
\textbf{Decision variables} &  \\
\(X_i^t \in \{0, 1\}\)& put container \(i\) on train \(t\)\\
\(B_i \in \{0, 1\}\)& put container \(i\) on truck\\
\textbf{Parameters}&  \\
\(e_i\)&  earliest day on which container \(i\) is available\\
\(l_i\)&  delivery due date for container \(i\) \\
\(d_t\)&  day on which train \(t\) departs the origin\\
\(ar_t\)&  day on which train \(t\) arrives at its destination\\
\(cap_t\)&  number of spaces available on train \(t\)\\
\(C_t\)&  costs of transporting a container with train \(t\)\\
\(C\) &  costs of transporting a container with a truck\\
\hline
\end{tabular}
\end{center}
\vspace{-0.5cm}
\end{table}

\begin{equation}\label{equ1}
\text{Minimize } \sum_{i\in I}\sum_{t\in T}{C_t} \cdot X_i^t + \sum_{i\in I}{C} \cdot {B_i}
\end{equation}

\textbf{Subject to:}

\begin{align}
\sum_{t\in T}X_i^t + {B_i} = 1&,& \forall i\in I \label{equ3}\\
X_i^t \cdot {d_t}\ge X_i^t \cdot {e_i}&,& \forall t\in T, i\in I \label{equ4}\\
X_i^t \cdot {ar_t}\ge X_i^t \cdot {l_i}&,& \forall t\in T, i\in I \label{equ5}\\
\sum_{i\in I}{X_i}^t \leq {cap_t}&,& \forall t\in T \label{equ6}
\end{align}

\subsection{Online Planning Problem Definition} \label{section3.2}

We represent the online multimodal transportation planning problem as a Markov Decision Process (MDP). In each state of the MDP, some containers have already been assigned and an action must be taken to assign the next container to a transportation resource, leading to a new state in which one more containers have been assigned. This must be done in such a way that the total (expected) costs of assigning containers to transportation resources is minimized. Consequently, the MDP is defined by the following elements.

The set of \textbf{states \(S\)}, where each state has two components. The first component of our states is a list of train capacities, \(\{cap_{1}, cap_{2}, \ldots, cap_{|T|}\}\), where each \(cap_{j}\) represents the number of slots available on the train or trains that correspond to a particular train schedule. In particular, \(cap_{1}\) represents the number of slots available on the train or trains with (departure day, arrival day at destination) equal to (1,1), \(cap_{2}\) is the total capacity available for (departure day, arrival day at destination) equal to (1,2), and so on. As a consequence, the arrival and departure times of trains are implicitly encoded in the state. Note that, if we have multiple trains with the same departure and arrival day, in this way they are part of the same $cap_j$.
The second component of our states is the information about the next container that must be assigned. More precisely, a container $i$ is represented by the earliest day on which this container is available at the logistics company \(e_i\) and the due delivery day  \(l_i\). The next container to assign is selected using a heuristic. Two alternative heuristics are compared as  explained further on in this paper.

The set of \textbf{actions \(A\)}, consisting of all possible train options \(T\) and an option `Truck' that is assumed to be always available and uncapacitated. Once an action \(a\) is chosen in state \(s\), the next state \(s'\) is determined by reducing the capacity of the selected train by 1, if a train is chosen, and by selecting the next container to plan. Note that not all actions are possible in each state, because of the constraints that apply (see Section~\ref{section3.1}). For example, a train could have no more slots available, or its scheduled departure time could not meet the due delivery date of the container.

The \textbf{reward function \(R (s, a)\)}, which is the negative cost associated with selecting an action \(a\) (transportation cost of the selected train/truck) from the list of eligible actions:

\begin{equation}\label{equ 7}
{
R(s, a)=
    \begin{cases}
      -C_{a}, \text{if $a\in T$}\\
      -C, \text{if $a = Truck$}
    \end{cases}
    }
\end{equation}

The \textbf{objective}, which is maximizing the expected cumulative reward of the selected actions. Note that this is equal to minimizing the expected cumulative cost of transportation. We use the Bellman equation~\cite{Barron1989} to calculate this.

\begin{algorithm}
\caption{Deep Q-Learning for Online Multimodal Transportation Planning}
\begin{algorithmic}[1]
\STATE Initialize Deep Q-Network $Q$
\STATE Initialize replay memory \(D\)
\FOR{episode = 1 to E}
\STATE Generate new containers and trains \label{alg:generate}
\STATE Set current state $s$ with random capacity for all trains
\WHILE{there is an unassigned container $i \in I$}
\STATE $A' \leftarrow mask(s)$ forbidden actions (Equation \ref{Aprime}) \label{alg:masking}
\STATE With probability \(\varepsilon\) select a random action \(a \in A'\) \label{alg:egreedy1}
\STATE Otherwise select \( a = argmax_{a'\in A'}Q(s, a')\) \label{alg:egreedy2}
\STATE Create new state $s'$ from $s$ by updating train capacity used by $a$ \label{alg:update_capacity}
\STATE Update new state $s'$ with next container \label{alg:new_arrival}
\STATE Calculate reward $r = R(s,a)$ \label{alg:reward}
\STATE Record experience $(s, a, r, s')$ in replay memory \(D\) \label{line12}
\STATE $s \leftarrow s'$

\IF{every $M$ iterations} \label{alg:startupdate}
\STATE Sample random minibatch of experience from replay memory \(D\)
\FOR{$(s, a, r, s')$ in minibatch}
\STATE $y \leftarrow$ Bellman Equation over $(s, a, r, s'), Q$
\STATE Update Deep Q-Network $Q(s, a)=y$ \label{line20}
\ENDFOR
\ENDIF
\ENDWHILE
\ENDFOR
\end{algorithmic}\label{algo1}
\end{algorithm}
\section{Deep Reinforcement Learning for Solving Online Planning Problem}\label{section4}
As discussed in Section \ref{section2}, Deep Reinforcement Learning proved to be able to tackle challenging problems in several industrial applications with promising results.
Therefore, in this paper we propose an algorithm based on Deep Q-Learning~\cite{Mnih2013} to solve the multimodal online planning problem introduced in previous section. Algorithm~\ref{algo1} summarizes our Deep Q-learning method. We discuss the various components more in details in the following subsections.

\subsection{Multimodal Transportation Problem Environment}\label{6.1}
The DRL algorithm learns by performing a number of episodes $E$. During each episode a set of containers is planned either on a train or on a truck. The \emph{environment} has the information on the trains and containers. It keeps a current state, and can be given actions to perform that will result in a reward and a new state (see Section~\ref{section3.2}). To this end, the environment has two main functions:
\begin{itemize}
    \item \textbf{Environment initialization}. At the beginning of each episode a new environment is generated by launching the data generator, to ensure that the starting point of each new episode is different from other episodes. The data generator creates: a set of trains with their temporal features and initial capacities, a set of containers, with their temporal features, and transportation costs for each vehicle option (Algorithm line~\ref{alg:generate}).
    \item \textbf{Interaction with the agent}. For training the Deep Reinforcement learning model, we need to have interaction between the agent, model, and environment. In this interaction, we update the environment, calculate the next state and calculate reward of this action. In our problem, updating the environment means updating the capacity of trains based on the selected action (Algorithm line~\ref{alg:update_capacity}).
        Then, a new state is generated using the updated train capacities from the environment, and selecting the next container to plan (Algorithm line~\ref{alg:new_arrival}). The reward for the selected action (see Section \ref{selection4.3}) by our agent is the negative cost of the transportation associated with the selected action (Algorithm line~\ref{alg:reward}). We compare two different heuristics for selecting the next container to assign, (1) Earliest arrival first (or First In First Out - FIFO) and (2) Earliest due date first (EDF). Hereafter, we refer to the method implementing the FIFO and the EDF allocation heuristic as ``DRL-FIFO''and ``DRL-EDF'', respectively.
\end{itemize}
\subsection{Feature Engineering and Deep Q-Network Architecture}
The algorithm learns through a Deep Q-Network, which learns the Q values for state/action combinations.

\begin{figure}

\begin{center}

\includegraphics[scale=0.27]{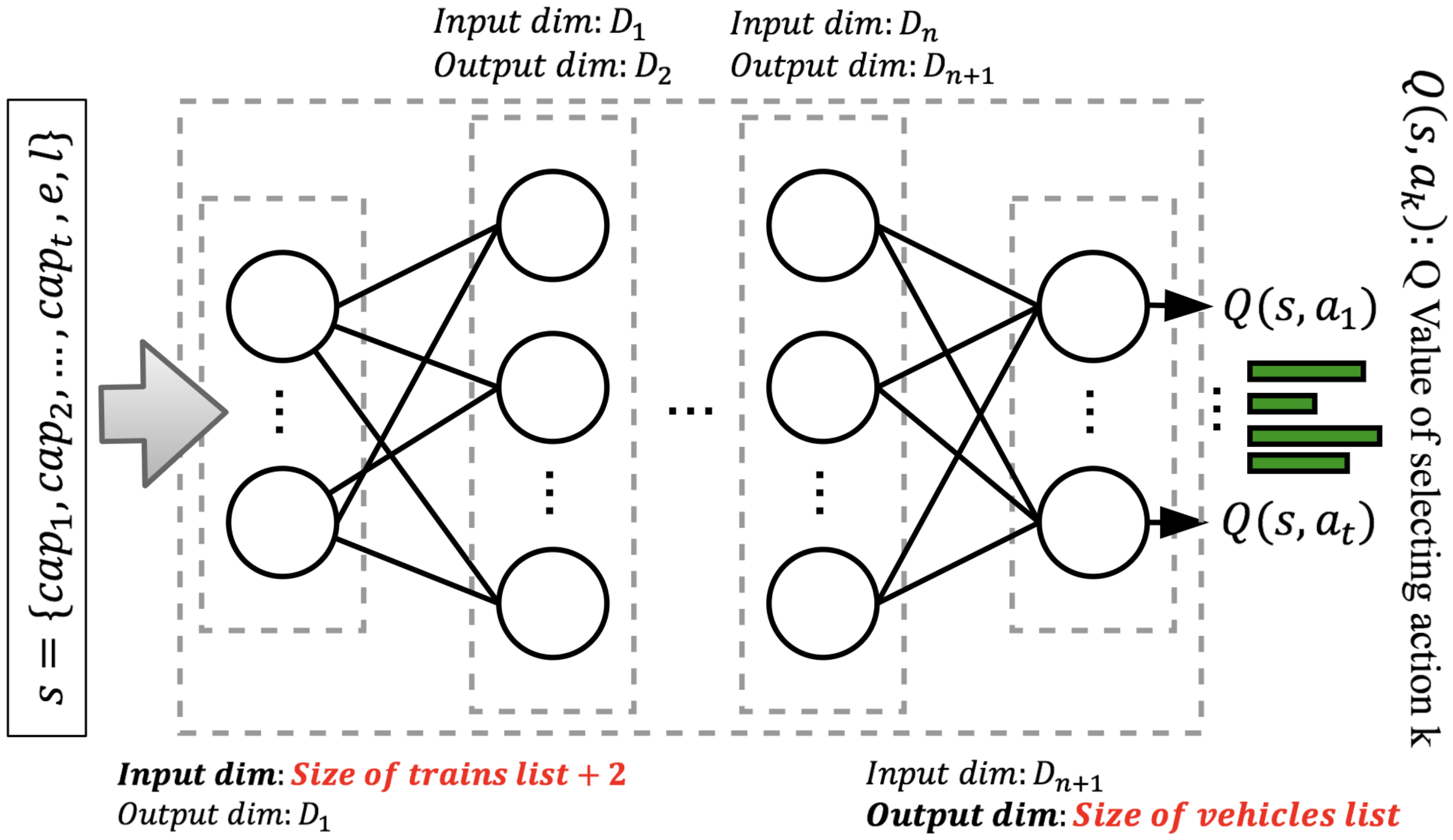}
\caption{Deep Q-Network Architecture} \label{network}
\end{center}
\vspace{-0.8cm}
\end{figure}

As explained in Section~\ref{section3.2}, we use as input features of the network a vector of size \(|T| + 2\), which consists of a list of the number of spaces available on trains and both temporal features \(e_i, l_i\) of container $i$. Accordingly, we have \(|T| + 2\) input nodes for the network. As output nodes, we use a separate output unit for each possible action. Hence, the size of our output layer is equal to the size of the vehicle options (\(A\)). The outputs of our Deep Q-Network correspond to the predicted Q-values of the individual action \(a\) for the input state \(s\).
Fig.~\ref{network} shows the overall neural network architecture, that is a fully-connected neural network with \(k\) hidden layers.

\subsection{Action Selection Methods and Masking Approach} \label{selection4.3}
In the transportation planning problem eligibility of actions changes dynamically (see  Section \ref{section3.2}), with the result that the list of allowed actions can be different for each state \(s\). 
However, the use of a dynamic set of actions increases significantly the complexity of the problem, up to the point where the computation is not feasible. 
To deal with this challenge, we determine a static actions list of all possible actions 
and then use a masking approach to determine which actions are enabled at each state \(s\).

To select an action to perform in a state, we use a customized epsilon-greedy method with masking. More precisely, in a state we first determine the set of possible actions through masking (Algorithm line~\ref{alg:masking}) as follows:
\begin{equation}\label{Aprime}
\footnotesize
mask(\{cap_{1}, \ldots, cap_{|T|},e,l\}) = \{ t \in T \mid d_{t}\ge e, ar_{t}\leq l, cap_{t}\ge 1\} \cup \{Truck\}
\end{equation}
We then apply the epsilon-greedy method to the eligible actions \(A'\) (Algorithm lines~\ref{alg:egreedy1}-\ref{alg:egreedy2}). In this method, the agent selects a random eligible action with a fixed probability, \(0\ge \varepsilon \ge 1\), or the action that is optimal with respect to the learned Q-function otherwise~\cite{Tokic2011}.

\subsection{Replay Memory and Minibatch}\label{selection4.4}
We use a replay memory~\cite{Mnih2013}. In this method, we record the experiences of our agent into a replay memory \(D\) at each step $(s, a, r, s')$ of each episode (Algorithm line \ref{line12}). Every $M$ steps, we then update the Deep Q-Network with the new experiences. The main advantage of this method consists in decreasing the variance of the updates. Lines~\ref{alg:startupdate} to \ref{line20} show how we apply Q-learning updates, or minibatch updates, by first sampling experiences randomly from the replay memory, calculating the expected cumulative reward for each experience using the Bellman equation and then updating the Deep Q-Network for each experience with the calculated expected cumulative reward.

\section{Experiments and Results}\label{section5}
In this section, we discuss the results of the experiments that we carried out to test the performance of our proposed methods. In Section \ref{section5.1}, we explain the dataset, the hyperparameter training setup, and the methods we tested in comparison to our approach. In Section \ref{section5.2}, we discuss the training and stability analysis.
In Section \ref{section5.3}, we report the performance of our method, tested with both the discussed container selection heuristics, and the other competitor planning methods we tested, together with an analysis on the optimality gap between offline and online methods.

\subsection{Experimental Settings}\label{section5.1}
\paragraph{Dataset} For this experiment, we generated data with properties that are based on the long-haul transportation planning problem of a logistics company for a particular transportation corridor (i.e., the set of available transportation resources between two particular transshipment points). As discussed in Section \ref{section4}, these data include the following features: the number of trains, the capacity and temporal properties of the trains, containers with their temporal features, and transportation costs. Time windows of this experiment are weekly. We assume that trucks are always available and the number of trucks is uncapacitated. This is in line with the experience of the company that they can always find charter companies to transport containers by truck.
\paragraph{Training parameters} We did hyperparameter tuning on: the number of episodes (with options 1000, 2000, 3000, 4000), learning rate (0.01, 0.1), number of hidden layers (2, 4), discount factor (0.5, 0.99) , number of nodes per hidden layer (100, 150, 200), and mini batch size (5, 10, 15).
The algorithm worked best and learning converged using \(E=4,000\) episodes of 7 days. Note that the starting state of each episode is different from other episodes. In each episode 100 containers must be planned, i.e. 100 steps must be performed. The number of containers is chosen proportional to the train capacity over the week, in line with the properties of the planning problem at the logistics company. Each container has an earliest availability day that is uniformly distributed over the week. The due date is uniformly distributed over the days after the earliest availability day.
There are 28 train schedules per week. 
For the capacity of trains in each train schedule we test 7 different settings, i.e.: 6 different settings in which each train schedule $(1,1), (1,2), \ldots$ has the same capacity 1 through to 6; and one setting in which each schedule has a random number of available slots that is uniformly distributed over 0 to 6 spaces. The goal of using these different capacity settings is to investigate the effect of the available capacity on the planners' performance.
We initialize a fully-connected feedforward neural network with backpropagation with 2 hidden layers of 100 nodes, ReLU activator, and Adam optimizer. We use a replay memory of size 10,000 and retrain the Deep Q-Network based on minibatches 5 times per epoch.
The discount factor, used in the Bellman equation, is \(\gamma =0.99\). This means that we care more about the reward that is received in the future than the immediate reward. Remaining parameters are initialized according to PyTorch’s default parameters. The probability $\epsilon$ with which a random action is chosen starts at 0.95 and is decreased after each episode in steps of 0.1 until it reaches 0.05.
The agent and the simulation model are executed on a machine with an Intel(R) Core(TM) i7 Processor CPU @ 2.80GHz and 16GB of RAM, no graphics module is used for training the neural network.

\paragraph{Planning Methods used for Comparison}

We compare the performance of our method against the results obtained by two groups of methods: (1) ILP-based (re) optimization, and (2) Greedy heuristics. These methods are inspired by the existing literature~\cite{VanRiessen2016} and discussions with the logistics company on how their online planner currently works. 
For the ILP-based planning, we run an ILP planner which returns a (sub) optimal, offline solution on the basis of the information known at a given moment in time. More precisely, we tested three different ILP-based planning methods. The first one, the \textit{ILP} solver, is run once per week and has perfect knowledge on all the containers for that week, as defined in Section~\ref{section3.1}. Consequently, this planner always produces the optimal solution. However, this solution is purely theoretical and only used for benchmarking, because the assumption that the precise arrivals of containers are all known at the start of the week is unrealistic. The \textit{2-ILP} planner is run twice per week, in both cases only with the information on the containers that have arrived up until and including the day of the planning. Finally, the \textit{7-ILP} planner is run daily, and also has information on containers that have arrived up until and including the day of the planning. While the ILP-solver computes the optimal solution for the offline planning problem, the 2-ILP and the 7-ILP only have limited information. These types of planners are commonly used in practice.
The greedy heuristics, entitled \textit{First train} and \textit{Cheapest train}, assign each container separately to an eligible train that will take it to the destination on time and, if no such train is available, to a truck. 
The first greedy heuristic assigns a container to the first available train. The second one assigns a container to the cheapest available train. 
The comparison is based on the total cost of transportation over 200 weeks, i.e. 20,000 containers.

\begin{figure*}
\begin{subfigure}[b]{0.33\linewidth}
  \centering
  \includegraphics[scale=0.3]{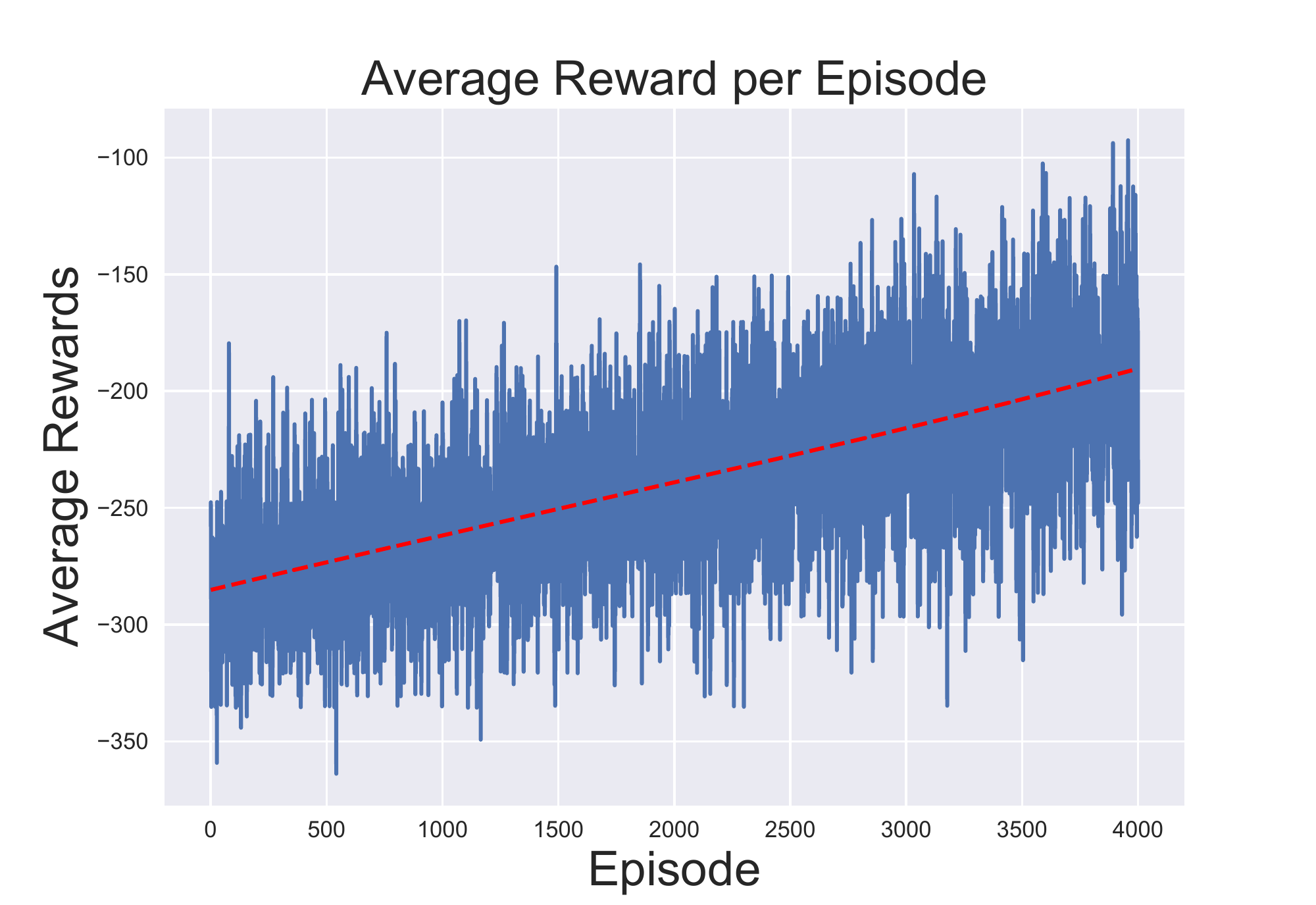}
  \caption{Average reward - DRL-FIFO}
  \label{fig:sfig1}
  \vspace{+0.35cm}
\end{subfigure}%
\begin{subfigure}[b]{0.33\linewidth}
  \centering
  \includegraphics[scale=0.3]{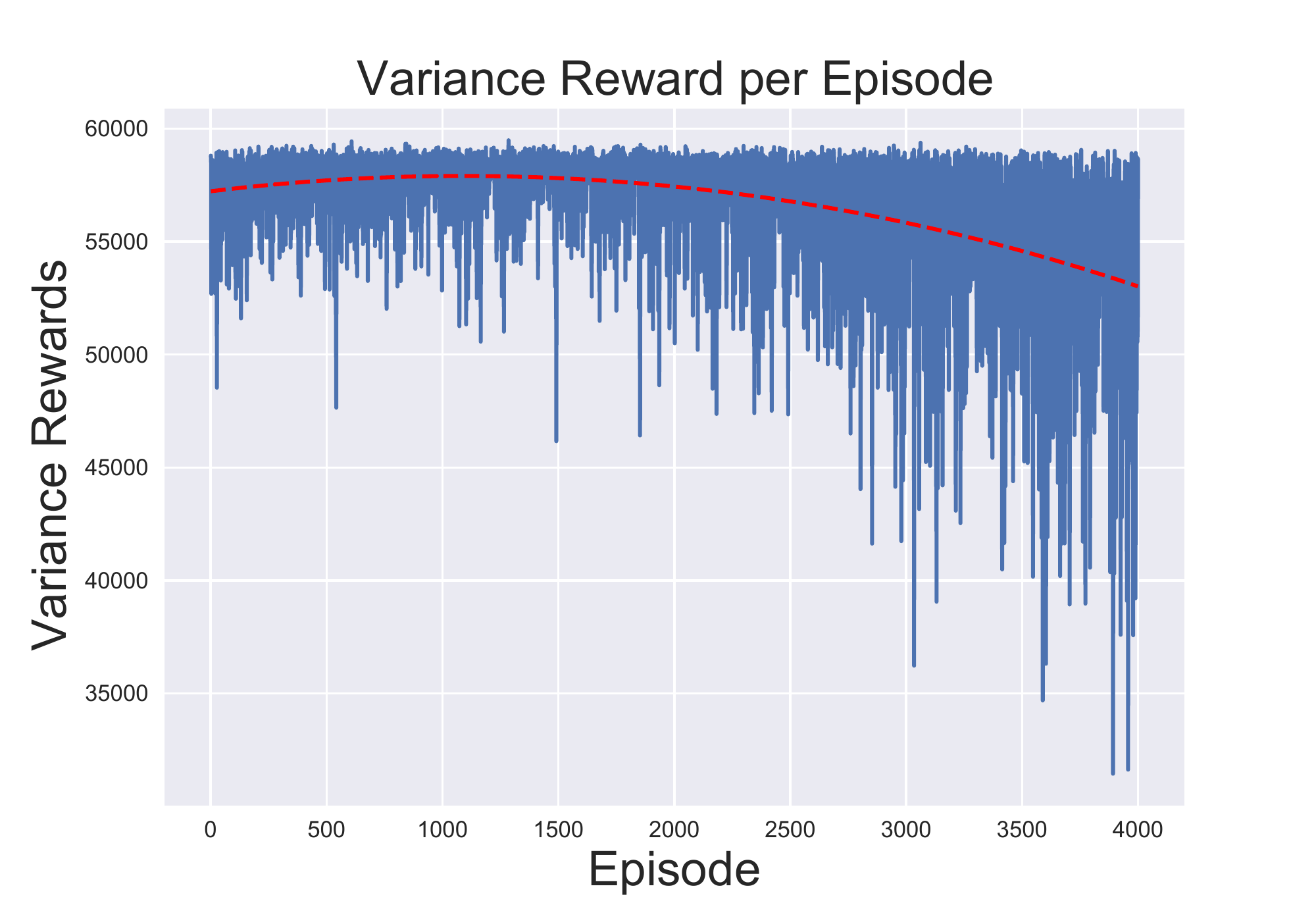}
  \caption{Variance reward - DRL- FIFO}
  \label{fig:sfig2}
   \vspace{+0.35cm}
\end{subfigure}
\begin{subfigure}[b]{0.33\linewidth}
  \centering
  \includegraphics[scale=0.3]{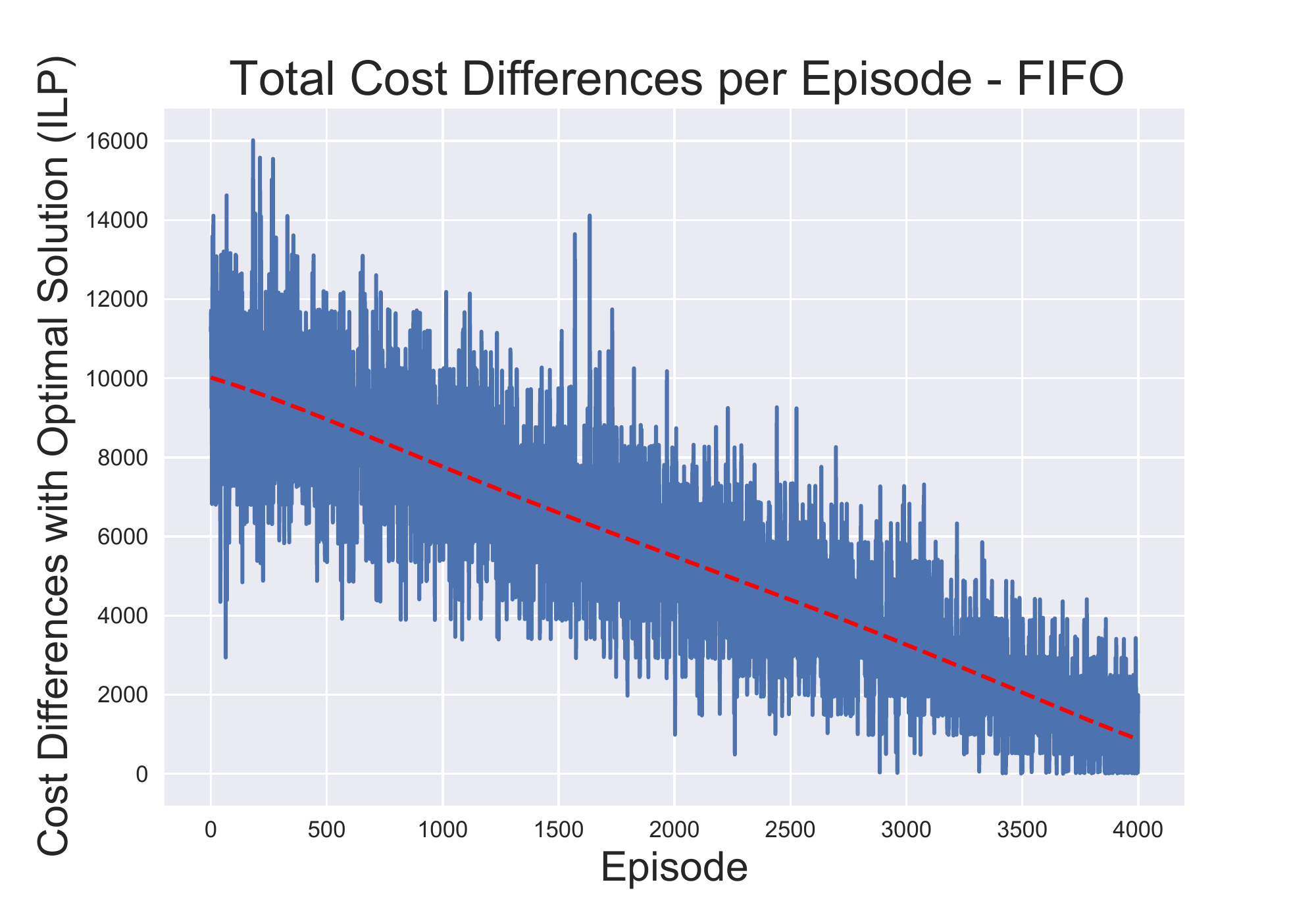}
  \caption{Cost differences with optimal solution- DRL-FIFO}
  \label{fig:sfig3}
\end{subfigure}\hfill
\begin{subfigure}[b]{0.33\linewidth}
  \centering
  \includegraphics[scale=0.3]{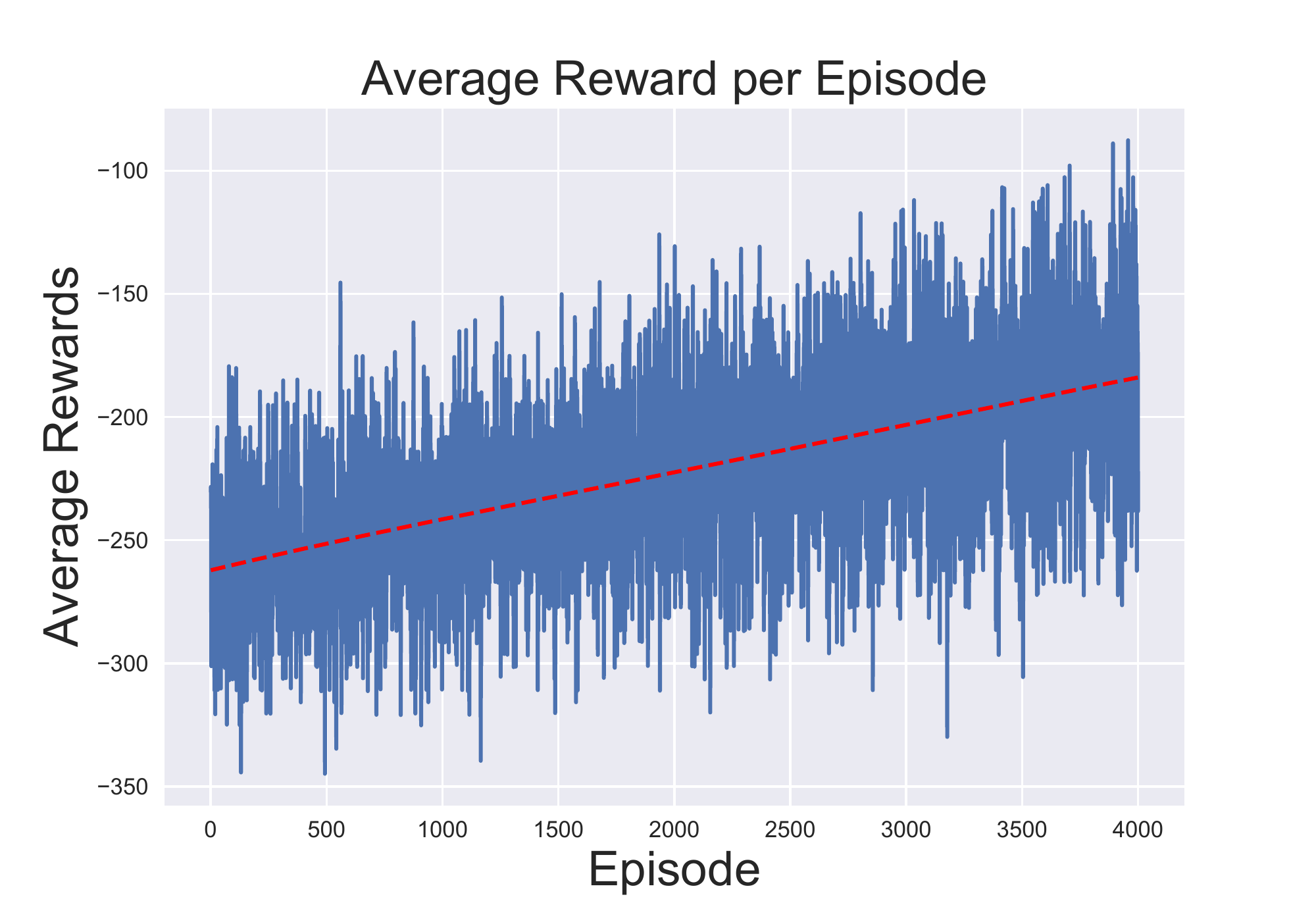}
  \caption{Average reward - DRL-EDF}
  \label{fig:sfig4}
  \vspace{+0.35cm}
\end{subfigure}%
\begin{subfigure}[b]{0.33\linewidth}
  \centering
  \includegraphics[scale=0.3]{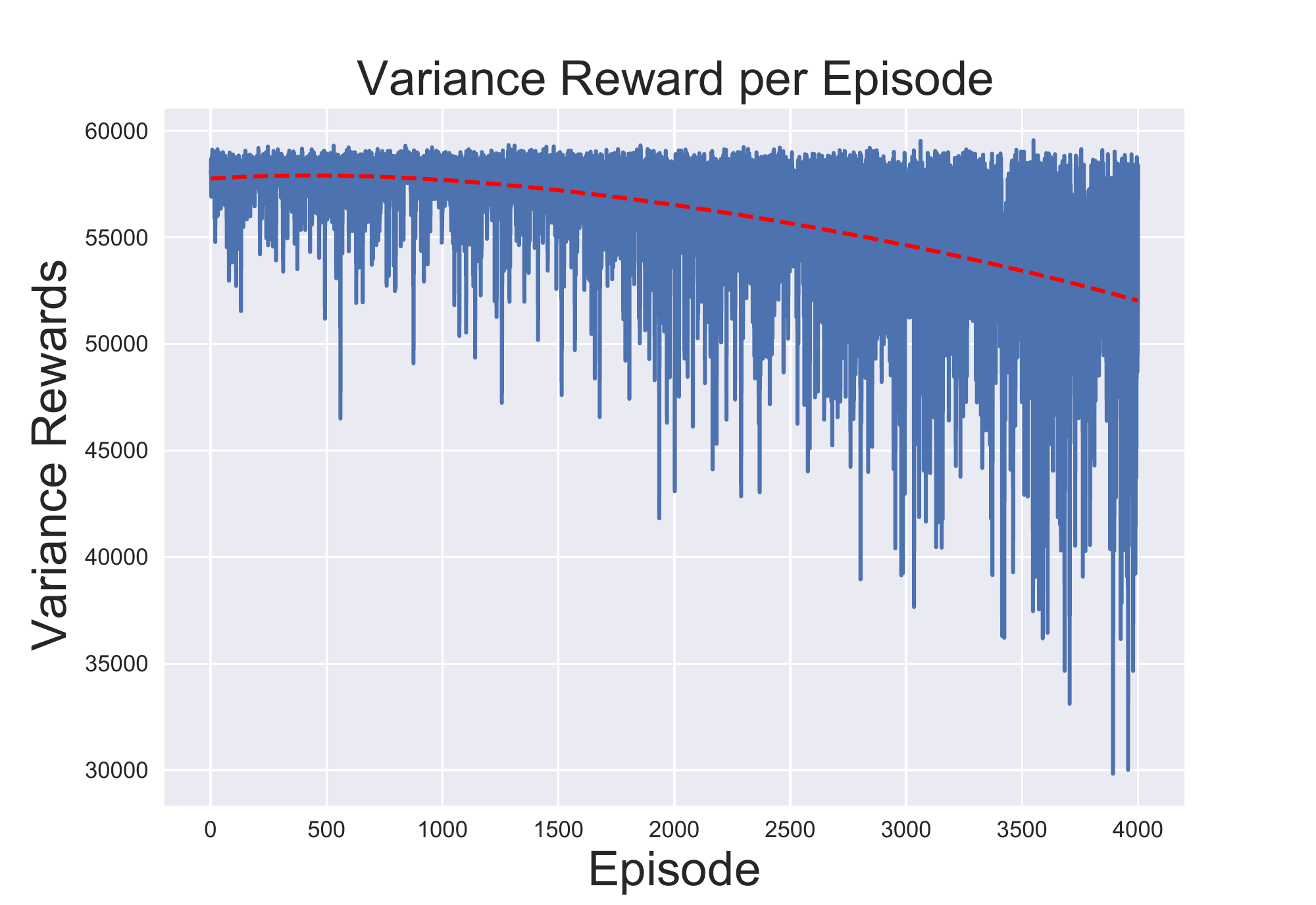}
  \caption{Variance reward - DRL-EDF}
  \label{fig:sfig5}
  \vspace{+0.35cm}
\end{subfigure}
\begin{subfigure}[b]{0.33\linewidth}
  \centering
  \includegraphics[scale=0.3]{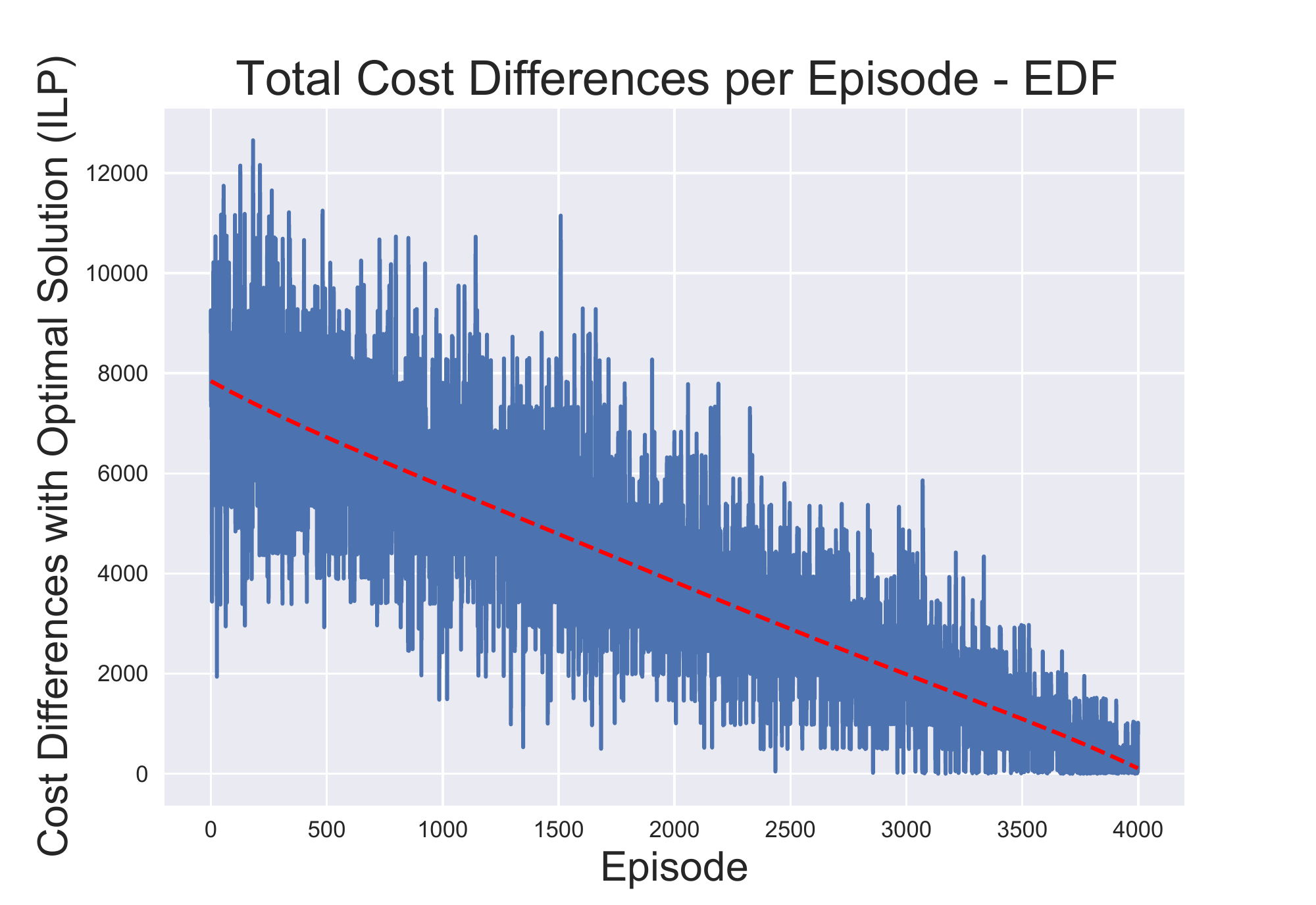}
  \caption{Cost differences with optimal solution- DRL-EDF}
  \label{fig:sfig6}
\end{subfigure}

\caption{ Plots of Average Reward, Variance Reward and Cost differences with optimal solution for DRL-FIFO anf DRL-EDF in the random capacity setting.
}
\label{DRL_train}
\vspace{-0.4cm}
\end{figure*}

\subsection{Training and Stability Analysis}\label{section5.2}
One of the popular evaluation measures for Deep Reinforcement Learning models is the total reward the agent collects in each episode during training~\cite{Bellemare2013}. In this subsection, we discuss this evaluation measure for the proposed DRL method with both the allocation heuristics discussed in subsection \ref{6.1}.
For the sake of space, we plot only the results of the setting with random capacity for each train. We chose this setting because it best reflects the reality, where trains usually do not have a fixed capacity. However, we would like to point out that the results obtained for all the other tested settings show the same trends as the setting selected for the discussion.
The plots in Fig.~\ref{DRL_train} illustrate the changes in the average/variance reward per episode 
and in the cost differences with the ILP solver per episode for both DRL-FIFO and DRL-EDF 
in the 4,000 episodes of the training. The red line is a regression line highlighting the behavior of the model during these 4,000 episodes. The figures show for both the algorithms a smooth improvement of the average rewards, as well as a visible decrease of the variance reward per episode. These results demonstrate that in both cases we did not experience any divergence issues in this learning process.
Fig.~\ref{fig:sfig3}, ~\ref{fig:sfig6} show that the cost differences between our method and the optimal solution in both cases converge to zero. These results show that our agent is able to learn container assignment patterns, getting closer and closer to the optimal solution as the training goes on. Furthermore,  
from the comparison of Fig.~\ref{fig:sfig3}, ~\ref{fig:sfig6}, we recognize learning allocation rules in DRL-EDF is faster than in DRL-FIFO.

\subsection{Methods Comparison and Optimality Evaluation }\label{section5.3}

\begin{figure}
\begin{center}
\includegraphics[scale=0.32]{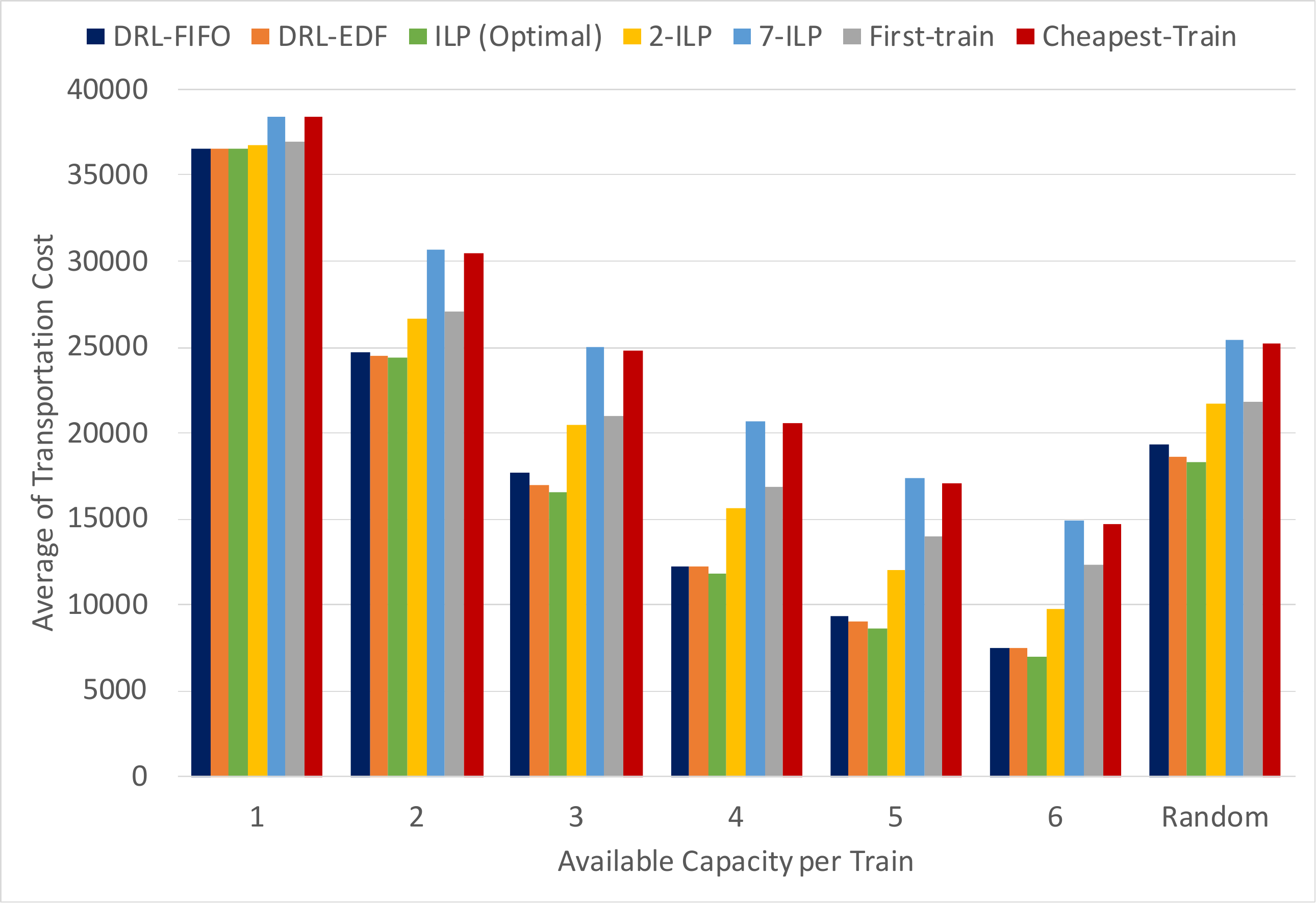}
\caption{Average transportation cost of different methods over 200 weeks.} \label{evaluation-cost}
\end{center}
\vspace{-0.6cm}
\end{figure}

\begin{figure}
\begin{center}
\includegraphics[scale=0.32]{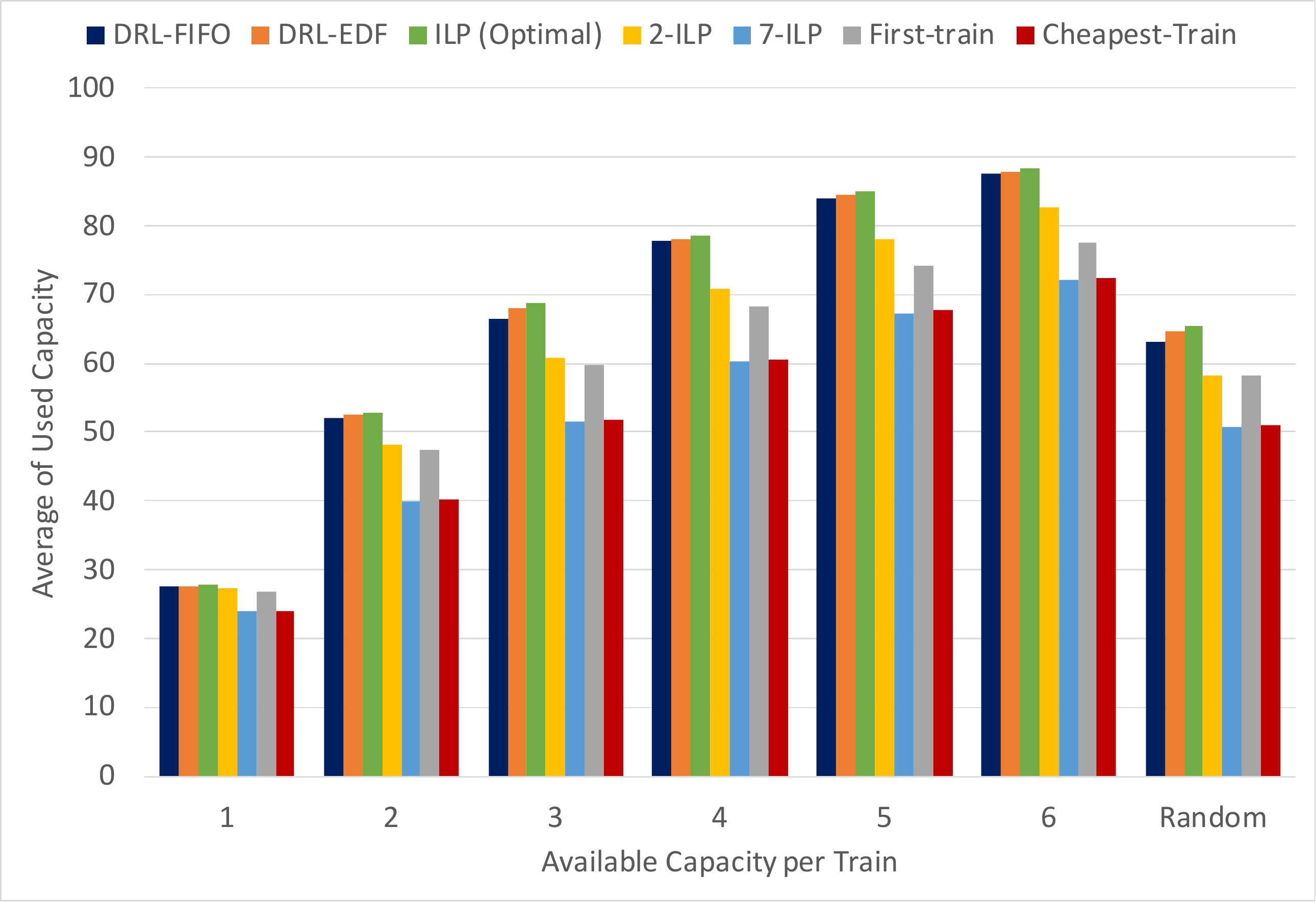}
\caption{Average capacity utilization of different methods over 200 weeks.} \label{evaluation-capacity}
\end{center}
\vspace{-0.5cm}
\end{figure}

In this subsection, we compare the performance of  DRL-EDF and DRL-FIFO with the ILP-based optimization methods and the greedy heuristics we introduced before.
The evaluation is performed based on the average 
of the total cost of transportation and utilization of capacity, that are computed respectively as the sum of all transportation costs of trains and trucks, and as the used train slots over each day. We tested these methods in different experiments with seven different capacity settings. Fig.~\ref{evaluation-cost} shows the distribution of the transportation costs and Fig.~\ref{evaluation-capacity} shows the distribution of the capacity utilization of each method using histogram in the different capacity settings. 
The costs/utilization are computed as the average per week over 200 weeks. 
 Fig.~\ref{evaluation-cost} and Fig.~\ref{evaluation-capacity} show that in the most competitive scenario (i.e., with train capacity equal to 1) the tested methods mostly obtain similar results, even though 7-ILP and the greedy methods perform worse both in terms of total costs and train utilization. However, in all the remaining capacity settings, DRL-FIFO and DRL-EDF consistently outperform 2-ILP, 7-ILP  and the greedy heuristics, obtaining much lower transportation costs and higher utilization of capacities. These differences are more evident with the increasing of the available capacity. Furthermore, the performance of DRL-FIFO and DRL-EDF are in all the tested settings very close to that of the optimal solution generated by the ILP solver.
These results also show that the use of ILP-based methods with limited knowledge (i.e., 2-ILP and 7-ILP) leads in general to poor performance, in some cases comparable or even worse than the results obtained using much simpler greedy methods. This is especially true for the 7-ILP planner, which demonstrates that daily optimum plans are often far from the global (weekly) optimum plans. These results are in line with the expectations. Indeed, planning daily and without taking into account containers to be scheduled in the following days resulted in several cases in containers having to be assigned to trucks, since trains which would have been suitable for their deadlines were already assigned to containers with earlier arrival day. This effect is less visible, but still relevant, for the 2-ILP planner, since it has more information available. Our DRL methods  outperform the offline methods with limited knowledge because they can learn the best container-train patterns, thus limiting the impact of having limited knowledge available at the moment of the decision. It is also interesting to note that the `First train' method mostly performs better than the other greedy method and than the 7-ILP planner. One of the reasons is that choosing the first available trains  may arise to increase available train options for upcoming containers.


Table.~\ref{tab3} reports differences of the average of the costs and capacity utilization of the different methods with respect to the optimal solution of the ILP solver, tested in different capacity settings.
Both DRL-FIFO and DRL-EDF  obtain on average values very close to the optimal solver. DRL-EDF obtained the best results, worsening the usage of capacity only of 0.71\% (against the 1.62\% of the DRL-FIFO) and the overall costs of 2.73\% (against the 4.70\% of DRL-FIFO). These results are encouraging, as ILP is offline and has access to the complete list of the containers, while our method works online for each container separately. We see that our learned policies are not too far from offline optimal solution, even though our online method never trains on instances with more than 100 containers per week. All the other methods show a significant worsening of the performance with respect to the optimal. For example, the 2-ILP method, which is the best one among the ILP-based re-optimization and the greedy algorithms, still obtains on average a capacity utilization lower than the optimal one of 8.23\%, as well as average increasing of the total costs of the 23.21\%.

\begin{table}
\begin{center}
\caption{Average differences with optimal solution over 200 weeks.}\label{tab3}
\setlength\tabcolsep{4pt}
  \begin{tabular}{lll}
    \toprule
    \text{Method} & \text{Capacity utilization ($\%$)} &
      \text{Total cost ($\%$)}\\
      \midrule
    DRL-FIFO & -1.62 & +4.70 \\
    DRL-EDF & -0.72 & +2.73 \\
    2-ILP & -8.23 & +23.21\\
    7-ILP & -21.26 & +58.32\\
    First train & -10.82 & +34.08\\
    Cheapest train & -20.81 & +56.95 \\
    
    \bottomrule
  \end{tabular}
\end{center}
  \vspace{-0.9cm}
\end{table}

\section{Conclusions and Future Work}\label{section6}

This work investigated the application of Deep Reinforcement Learning in solving a challenging online planning problem in the multimodal transportation domain: the optimal assignment of containers to onward transportation, taking into account time, capacity constraints, and optimizing the total cost of transportation over all containers. We formulated the problem as a Markov Decision Process and, based on this formulation, we developed a DRL algorithm using two different allocation heuristics able to carry out online planning of single containers, with the goal of minimizing the overall transportation costs.
The approach has been tested using data simulating a realistic scenario, designed on the basis of a real case study from a logistics company. The experimental results revealed that the proposed method is able to learn patterns of containers assignment in our scenario and the performance of the DRL-EDF is better than the one of DRL-FIFO.
We have compared the performance of both DRL-EDF and DRL-FIFO against the results of two ILP based re-optimization methods and two greedy heuristics commonly used for online planning, as well as against the optimal ILP solution.
The results show 
that the proposed DRL method outperformed the tested competitors in terms of total transportation costs and utilization of train capacity by 20.48\% to 55.32\% for the cost and by 7.51\% to 20.54\% for the capacity. Furthermore, it obtained
results within 2.7\% for the cost and 0.72\% for the capacity of the optimal solution generated by an ILP solver in an offline setting.
Overall, these results show how the use of AI-driven planners using Deep Reinforcement Learning can significantly decrease costs associated to container replanning for logistics companies, thus suggesting that the use of these techniques can indeed bring significant practical advantages in the logistic domain.
Nevertheless, our method presents some limitations that we plan to address in future work. In particular, the current version of the method has been designed and tested to support the allocation of containers to a single vehicle, rather than to a combination of vehicles. 
Furthermore, here we considered only two main transportation modes, i.e., trains and trucks. In future work, we plan to extend our model to incorporate these aspects, thus increasing the generality of the methods. Also, we will want to take locations as a planning factor into account. Furthermore, we intend to integrate the uncertainty aspect in the online planner algorithm. In particular, we intend to investigate how to integrate knowledge about probabilities of vehicles delay in the planning decision-making, to make more effective and less costly allocation plans.

\begin{thebibliography}{99}


\bibitem{Ishfaq2012}
Ishfaq, R. , Sox, C. R.: Design of intermodal logistics networks with hub delays. European Journal of Operational Research \textbf{220}(3), 629--641 (2012)


\bibitem{Brands2014}
Brands, T.,  van Berkum, E. C., Wismans, L. J. J.: Multi-objective optimization of multimodal passenger transportation networks: coping with demand uncertainty. In Proc. of 1st International Conference on Engineering and Applied Sciences, OPT-I, NTUA Press, 547--561 (2014)


\bibitem{VanRiessen2016}
Van Riessen, B., Negenborn, R. R., Dekker R.: Real-time container transport planning with decision trees based on offline obtained optimal solutions. Decision Support Systems \textbf{89}, 1--16 (2016)



\bibitem{Demir2015}
Demir, E., Huang, Y., Scholts, S., Van Woensel, T.: A selected review on the negative externalities of the freight transportation: Modeling and pricing. Transportation research part E: Logistics and transportation review \textbf{77}, 95--114 (2015)



\bibitem{Demir2016}
Demir, E., Burgholzer, W., Hrušovský, M., Arıkan, E., Jammernegg, W. and Van Woensel, T.: A green intermodal service network design problem with travel time uncertainty. Transportation Research Part B: Methodological \textbf{93}, 789--807 (2016)


\bibitem{Tian2017}
Tian, W., Cao, C.: A generalized interval fuzzy mixed integer programming model for a multimodal transportation problem under uncertainty. Engineering Optimization \textbf{49}(3), 481--498 (2017)


\bibitem{Yang2016}
Yang, K., Yang, L., Gao, Z.: Planning and optimization of intermodal hub and spoke network under mixed uncertainty.Transportation Research Part E: Logistics and Transportation Review \textbf{95}, 248--66 (2016)
 

\bibitem{Rivera2017}
Rivera, A. E. P. and Mes, M. R.: Anticipatory Scheduling of Freight in a Synchromodal Transportation Network. Transp. Res. Part E: Logist. Transp. Rev \textbf{105}, 176--94 (2017)


\bibitem{Sun2018}
Sun, Y., Hrušovský, M., Zhang, C., Lang, M.: A time-dependent fuzzy programming approach for the green multimodal routing problem with rail service capacity uncertainty and road traffic congestion. Complexity, (2018)


\bibitem{Sun2019}
Sun, Y., Li, X., Liang, X., Zhang, C.: A bi-objective fuzzy credibilistic chance-constrained programming approach for the hazardous materials road-rail multimodal routing problem under uncertainty and sustainability. Sustainability \textbf{11}(9), (2019) 



\bibitem{QU2019}
Qu, W., Rezaei, J., Maknoon, Y., Tavasszy, L.: Hinterland freight transportation replanning model under the framework of synchromodality. Transportation Research Part E: Logistics and Transportation Review \textbf{131}, 308--28 (2019)

 

\bibitem{Gumuskaya2020}
Gumuskaya, V., van Jaarsveld, W., Dijkman, R., Grefen, P., Veenstra, A.: Dynamic barge planning with stochastic container arrivals. Transportation Research Part E: Logistics and Transportation Review \textbf{}, (2020)



\bibitem{Nazari2018}
Nazari, M., Oroojlooy, A., Snyder, L., Taka, M.: Reinforcement learning for solving the vehicle routing problem. In Advances in Neural Information Processing Systems, 9839--9849 (2018)


\bibitem{Applegate2006}
Applegate, D. L., Bixby, R. E., Chvatal, V., Cook, W. J.: The traveling salesman problem: a computational study. Princeton university press, Princeton (2006)


\bibitem{Vinyals2017}
Vinyals, O., Fortunato, M., Jaitly, N.: Pointer networks. In Proceedings of the 29th Conference on Neural Information Processing Systems (NIPS), 2692--2700 (2015)


\bibitem{Bello2017}
Bello, I., Pham, H., Le, Q. V., Norouzi, M., Bengio, S.: Neural combinatorial optimization with reinforcement learning. In ICLR (Workshop), (2017)

\bibitem{Mnih2016}
Mnih, V., Badia, A. P., Mirza, M., Graves, A., Lillicrap, T., Harley, T., Silver, D., Kavukcuoglu, K.: Asynchronous methods for deep reinforcement learning. In International conference on machine learning, 1928--1937 (2016)

\bibitem{Khalil2017}
Khalil, E., Dai, H., Zhang, Y., Dilkina, B., Song, L.: Learning combinatorial optimization algorithms over graphs. In Proceedings of the 31st Conference on Neural Information Processing Systems (NIPS), 6348--6358 (2017)


\bibitem{Deudon2018}
Deudon, M., Cournut, P., Lacoste, A., Adulyasak, Y., Rousseau, L. M.: Learning heuristics for the tsp by policy gradient. In Proc. of CPAIOR, 170--181 (2018)


\bibitem{Kool2019}
Kool, W., van Hoof, H., Welling, M.: Attention, learn to solve routing problems! In Proceedings of the 7th International Conference on Learning Representations (ICLR), 2019.


\bibitem{daCosta2020}
da Costa, P. R. D. O., Rhuggenaath, J., Zhang, Y. and Akcay, A.: Learning 2-opt heuristics for the traveling salesman problem via deep reinforcement learning. In Asian Conference on Machine Learning, PMLR, 465--480 (2020).


\bibitem{Zhao2020}
Zhao, J., Mao M., Zhao, X., Zou, J.: A hybrid of deep reinforcement learning and local search for the vehicle routing problems. IEEE Transactions on Intelligent Transportation Systems, (2020)

\bibitem{Waschneck2018}
Waschneck, B., Reichstaller, A., Belzner, L., Altenmuller, T., Bauernhansl, T., Knapp, A., Kyek, A.: Deep reinforcement learning for semiconductor production scheduling. In 2018 29th Annual SEMI Advanced Semiconductor Manufacturing Conference (ASMC), 301--306 (2018)

\bibitem{Chen2019}
Chen, S., Fang, S., Tang, R.: A reinforcement learning based approach for multi-projects scheduling in cloud manufacturing. International Journal of Production Research \textbf{57}(10), 3080--3098 (2019)


\bibitem{Park2019}
Park, I. B., Huh, J., Kim, J., Park, J.: A Reinforcement Learning Approach to Robust Scheduling of Semiconductor Manufacturing Facilities. IEEE Transactions on Automation Science and Engineering, (2019)


\bibitem{Paraschos2020}
Paraschos, P. D., Koulinas, G. K., Koulouriotis, D. E.: Reinforcement learning for combined production-maintenance and quality control of a manufacturing system with deterioration failures. Journal of Manufacturing Systems \textbf{56}, 470--483 (2020)


\bibitem{Afshar2019-1}
Afshar, R. R.,  Zhang, Y., Firat, M., Kaymak, U.: A Reinforcement Learning Method to Select Ad Networks in Waterfall Strategy. BNAIC/BENELEARN, (2019)


\bibitem{Afshar2019-2}
Afshar, R. R.,  Zhang, Y., Firat, M., Kaymak, U.: Reinforcement Learning Method for Ad Networks Ordering in Real-Time Bidding. International Conference on Agents and Artificial Intelligence, 16--36 (2019)


\bibitem{Meisheri2020}
Meisheri, H., Baniwal, V., Sultana, N. N., Khadilkar, H., Ravindran, B.: Using Reinforcement Learning for a Large Variable-Dimensional Inventory Management Problem. In Adaptive Learning Agents Workshop at AAMAS, (2020)



\bibitem{Ghavamipoor2019}
Ghavamipoor, H., Golpayegani, S. A. H.: A Reinforcement Learning Based Model for Adaptive Service Quality Management in E-Commerce Websites. Business \& Information Systems Engineering \textbf{62}(2), 159--177 (2019)


\bibitem{Mnih2013}
Mnih, V., Kavukcuoglu, K., Silver, D., Graves, A., Antonoglou, I., Wierstra, D., Riedmiller, M.: Playing atari with deep reinforcement learning. arXiv preprint arXiv:1312.5602, (2013)


\bibitem{Tokic2011}
Tokic, M., Günther, P.: Value-difference based exploration: adaptive control between epsilon-greedy and softmax.  In Annual Conference on Artificial Intelligence, 335--346 (2011)


\bibitem{Bellemare2013}
Bellemare, M. G, Naddaf, Y., Veness, J., Bowling, B.: The arcade learning environment: An evaluation platform for general agents. Journal of Artificial Intelligence Research \textbf{47}, 253--279 (2013)



\bibitem{Barron1989}
Barron, E. N., Ishii, H.: he Bellman equation for minimizing the maximum cost. Nonlinear Analysis: Theory, Methods \& Applications \textbf{13}(9), 1067--1090 (1989)






\end{thebibliography}
\end{document}